\documentclass[sigconf]{acmart}

\usepackage{enumitem}
\usepackage{amsmath}
\usepackage{colortbl}

\newcommand{\us}{T{\small AB}C{\small F}}

\newcommand{\tabsyn}{TABSYN}
\newcommand{\wachter}{Wachter}
\newcommand{\dice}{DiCE}
\newcommand{\revise}{REVISE}
\newcommand{\cchvae}{CCHVAE}

\newcommand{\adult}{Adult}
\newcommand{\lending}{Lending Club}
\newcommand{\gmc}{Give me some credit}

\newcommand{\bank}{Bank marketing}
\newcommand{\default}{Credit default}

\usepackage{xcolor}
\usepackage{multirow}
\usepackage{makecell}

\definecolor{darkgreen}{rgb}{0.0, 0.5, 0.0}

\AtBeginDocument{%
  }

\copyrightyear{2024}
\acmYear{2024}
\setcopyright{rightsretained}
\acmConference[ICAIF '24]{5th ACM International Conference on AI in Finance}{November 14--17, 2024}{Brooklyn, NY, USA}
\acmBooktitle{5th ACM International Conference on AI in Finance (ICAIF '24), November 14--17, 2024, Brooklyn, NY, USA}
\acmDOI{10.1145/3677052.3698673}
\acmISBN{979-8-4007-1081-0/24/11}
\begin{document}

\title{TABCF: Counterfactual Explanations for Tabular Data Using a Transformer-Based VAE}

% \affiliation{%

\author{Emmanouil Panagiotou}
\authornote{Both authors contributed equally to this research.}
\affiliation{%
  \institution{Freie Universität Berlin}
  \city{Berlin}
  \country{Germany}}
\email{emmanouil.panagiotou@fu-berlin.de}

\author{Manuel Heurich}
\authornotemark[1]
\affiliation{%
  \institution{Freie Universität Berlin}
  \city{Berlin}
  \country{Germany}}
\email{manuel.heurich@fu-berlin.de}

\author{Tim Landgraf}
\affiliation{%
  \institution{Freie Universität Berlin}
  \city{Berlin}
  \country{Germany}}
\email{tim.landgraf@fu-berlin.de}

\author{Eirini Ntoutsi}
\affiliation{%
  \institution{Universität der Bundeswehr München}
  \city{Munich}
  \country{Germany}}
\email{eirini.ntoutsi@unibw.de}

\renewcommand{\shortauthors}{Panagiotou et al.}

\begin{abstract}
In the field of Explainable AI (XAI), counterfactual (CF) explanations are one prominent method to interpret a black-box model by suggesting changes to the input that would alter a prediction. In real-world applications, the input is predominantly in tabular form and comprised of mixed data types and complex feature interdependencies. These unique data characteristics are difficult to model, and we empirically show that they lead to bias towards specific feature types when generating CFs. To overcome this issue, we introduce \emph{\us}, a CF explanation method that leverages a transformer-based Variational Autoencoder (VAE) tailored for modeling tabular data. Our approach uses transformers to learn a continuous latent space and a novel Gumbel-Softmax detokenizer that enables precise categorical reconstruction while preserving end-to-end differentiability. Extensive quantitative evaluation on five financial datasets demonstrates that \us\ does not exhibit bias toward specific feature types, and outperforms existing methods in producing effective CFs that align with common CF desiderata. 
\end{abstract}





\maketitle

\section{Introduction}

Although Deep Neural Networks (DNN) are highly effective, their complexity makes them difficult to explain, hindering their adoption in crucial fields like healthcare and finance. Explainable AI (XAI) aims to overcome this by making their decisions interpretable \cite{adadi2018peeking}.

\begin{figure}[ht]
  \centering
  \includegraphics[width=\linewidth]{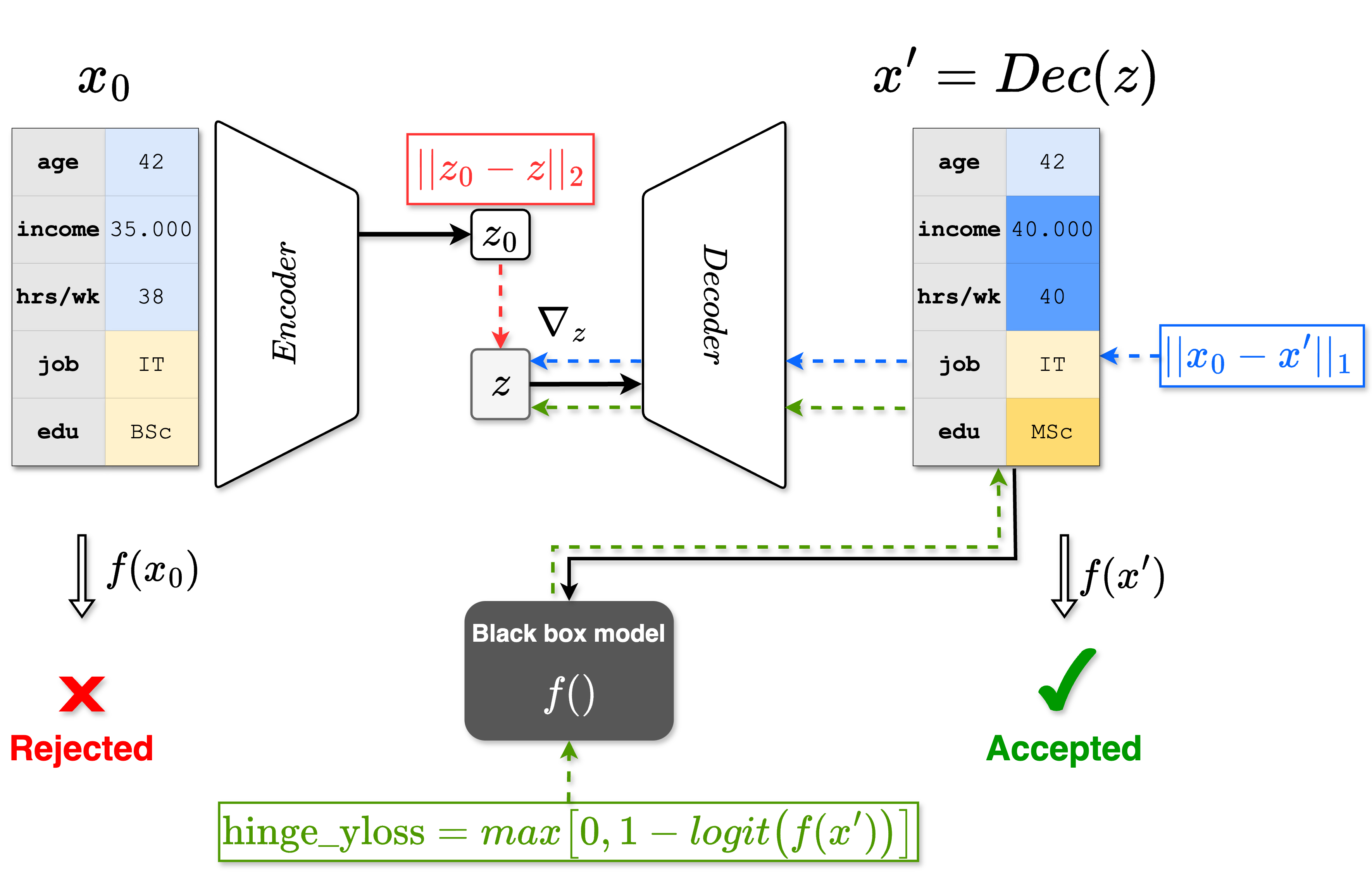}
  \caption{Overview of the counterfactual generation process. The bold arrows indicate data flow, and the dashed arrows indicate backward gradient flow. We iteratively optimize the latent representation $z$, of the counterfactual $x'$, using three distinct loss terms.}
  \label{fig:cf_optimization}
  \Description{Counterfactual generation process.}
\end{figure}

Counterfactual (CF) explanations are one way to gain insight into a black-box model's decision by offering meaningful changes to the input that would result in a favorable outcome. Most initial works find CFs by searching in the input space \cite{dice,wachter}. More recent studies leverage generative models, such as Variational Autoencoders \cite{vae} (VAE). These models capture the underlying structure of the data to generate more realistic and semantically meaningful CFs. However, most of these methods are developed for the vision domain \cite{hvilshoj2021ecinn,cem}, where generative models are particularly effective. In business applications, especially finance, data most often comes in tabular form presenting specific challenges like mixed (numerical and categorical) feature types, inherent imbalances, and complex feature interdependencies. Existing tabular CF methods handle the mixed feature space using simple pre/post-processing \cite{cruds,revise,clue,cchvae} or regularization functions \cite{dice}. Our experiments show that such data handling leads to \emph{feature-type bias}, for example predominantly changing numerical features when generating CFs.  

To address these challenges, we present \us, a CF generation method that employs a tranformer-based VAE tailored for tabular data. Our approach maps the mixed input into a unified continuous latent space, leveraging transformers to capture the rich feature interdependencies. Similar to other methods \cite{wachter, dice, revise} we define CF generation as an optimization problem, and use gradient descent to navigate the latent space. Importantly, our decoder architecture enables precise categorical reconstruction via the Gumbel-Softmax trick \cite{gumbel}, while being fully differentiable to ensure optimal gradient flow from the black-box model. Furthermore, we optimize for important CF desiderata \cite{verma2020counterfactual}, such as validity, but also, proximity to the original instance, and feature sparsity, for producing more \emph{actionable} CFs. We compare \us's performance against baselines in terms of producing valid and actionable CFs for tabular data, on binary classification problems. Our extensive quantitative evaluation on five financial and census datasets showcases that \us\ is superior in producing effective CF explanations, and does not exhibit feature-type bias. Our contributions include: i) identifying issues in the tabular data handling of existing methods, which impede optimization and result in feature-type bias, and ii) introducing \us\, a novel counterfactual generation method that employs a transformer-based VAE specifically designed for tabular data to address these limitations. 

Our paper is structured as follows. We describe all relevant works related to counterfactuals and generative models in Section~\ref{sec:related}, we introduce our method \us\ in Section~\ref{sec:method}, defining the transformer-based VAE and the CF generation process in the latent space. In Section~\ref{sec:experiments} we present our experimental evaluation, including datasets, metrics, and baselines, and in Section~\ref{sec:results} we present the results. We conclude the paper with a discussion and opportunities for future work in Section~\ref{sec:conclusion}. We have provided a codebase \footnote{\href{https://github.com/Panagiotou/TABCF}{github.com/Panagiotou/TABCF}} for reproducing all experiments.

\section{Related Work}
\label{sec:related}

\begin{figure*}[ht]
  \centering
  \includegraphics[width=\linewidth]{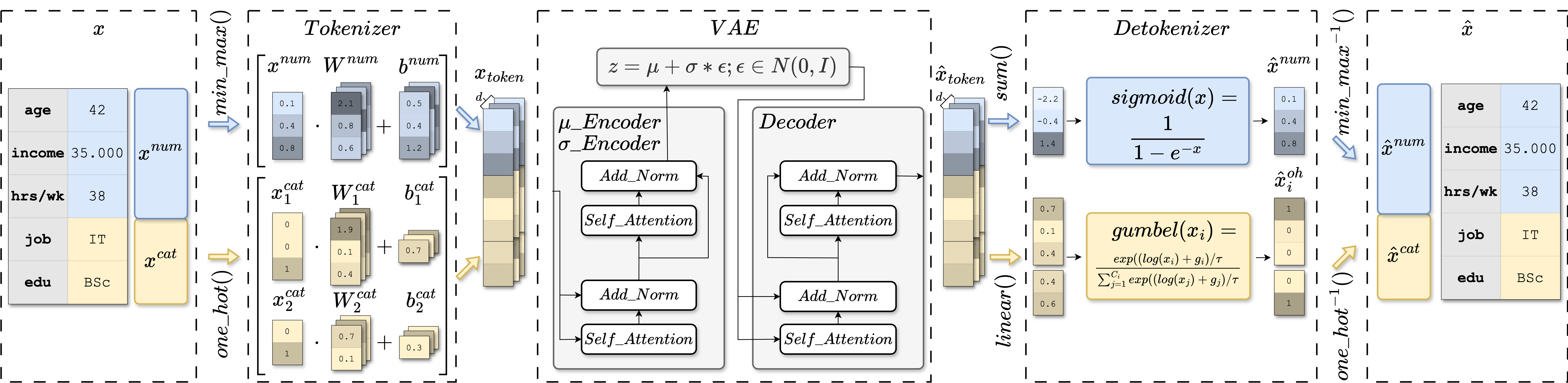}
  \caption{Overview of the Variational Autoencoder training process. Blue indicates the process for numerical, yellow for categorical features. The detokenizer enables a fully differentiable pipeline for categorical features using the Gumbel-Softmax function for sample reconstruction.}
  \label{fig:vae}
  \Description{VAE training process.}
\end{figure*}

% Our work is related to counterfactual explanation methods and generative models. The domain of XAI, particularly CF explanations, recently received significant attention due to the growing demand for interpretable AI. Consequently, 

Recent literature is abundant with studies on CFs that detail numerous desired properties (desiderata), target different data modalities, and employ various methodologies \cite{verma2020counterfactual} and criteria for evaluation \cite{carla}. This section provides an overview of the methods most pertinent to our field of work of tabular CFs.

\noindent{\textbf{Counterfactual desiderata.}} Are desirable properties for effective CF explanations. The most essential property is \emph{validity}, ensuring the proposed changes alter the decision of the black-box model we aim to explain. The vast majority of works simultaneously optimize to find the most minimal changes that lead to the desired outcome \cite{dice,wachter,revise,cchvae}. This objective involves the desiderata of \emph{proximity} and \emph{sparsity}, i.e., CFs that are close to the original instance and alter as few features as possible. Finally, some methods find more robust CFs that withstand minor perturbations \cite{bajaj2021robust}, account for predictive uncertainty \cite{clue}, causal constraints \cite{revise,naumann2021consequence}, and plausibility \cite{dandl2020multi}. 

\noindent{\textbf{Counterfactual methods.}}  Wachter et al. \cite{wachter} first introduced CF explanations as an optimization problem in input space, aimed at changing a model's decision while minimizing the distance to the original instance. This approach inspired various other methods that assume a differentiable black-box model and use gradient descent to find CFs, either in the input space like \dice\ \cite{dice} or in the latent space of a generative model like \revise\ and \cchvae\ \cite{revise,cchvae}. Our method falls within the latter category, and therefore we provide a detailed description of these methods in Section~\ref{sec:baselines}, and use them as baselines in our experiments. Other works assume a model agnostic setting where differentiability is not guaranteed, e.g. in decision trees. Most methods in this scenario use a heuristic search, like genetic algorithms \cite{naumann2021consequence}, or reinforcement learning to learn a policy for finding CFs \cite{verma2022amortized}.

\noindent{\textbf{Generative models.}} Several works rely on generative models to create CFs. While some methods \cite{del2024generating} use Generative Adversarial Networks (GANs) \cite{goodfellow2014generative}, the majority \cite{clue, cruds, cchvae, revise} prefer Variational Autoencoders (VAEs) \cite{vae}, for their flexibility and smooth latent representations. However, most generative models are designed for computer vision \cite{hvilshoj2021ecinn,dosovitskiy2020image} and are difficult to adapt to the unique characteristics of mixed tabular data \cite{ctgan}. Recently, transformers have emerged as a potential solution, due to their ability to capture complex feature interdependencies \cite{tokenizer}. Specifically, the \tabsyn\ \cite{tabsyn} method utilizes a transformer-based VAE for synthetic tabular data generation, outperforming existing approaches. To the best of our knowledge, our method \us\ is the first to use a transformer-based autoencoding framework for tabular counterfactual generation.

% Consequently, there are many classes of methods using various techniques to find CFs \cite{verma2020counterfactual}, from genetic algorithms \cite{naumann2021consequence}, gradient-based \cite{wachter}, to reinforcement learning \cite{panagiotou2023learning}. Our work is most related to methods that search for counterfactuals in input space or the latent space learned by some generative model. We have identified such state-of-the-art (SOTA) methods as competitors for our experimental evaluation, and use the implementations from the CARLA \cite{carla} framework.  

\section{\us: Counterfactual explanations for tabular data}
\label{sec:method}

In this section, we introduce our method \emph{\us}. We start with preliminaries, then we present the architecture of our transformer-based VAE in Section \ref{sec:architecture} and we describe the CF generation process in Section \ref{sec:generation}.

% \subsection{Problem statement - tabular counterfactuals}

We assume a differentiable black-box classifier $ f: \mathcal{X} \rightarrow \mathcal{Y}$ where $\mathcal{X}$ is the input feature space and $\mathcal{Y}=\{0, 1\}$ is the binary class. For an instance classified in the \emph{undesired class} $f(x) = 0$, the goal is to find a CF example that belongs to the \emph{target class}, i.e. $f(x') = 1$.  Additionally, we assume a mixed input space of $|N|$ numerical and $|C|$ categorical features $x=[x^{num}, x^{cat}] \in \mathbb{R}^{|N|+|C|}$. We train $f$ by pre-processing the data in the usual fashion so that all numerical features are min-max normalized, and all categorical features are one-hot encoded. Therefore, each row is presented as a $k$-dimensional vector $x=[x^{num}_1, x^{num}_2, \ldots, x^{num}_{|N|}, x^{oh}_1, x^{oh}_2, \ldots, x^{oh}_{|C|}]$, with $k = \mathbb{R}^{|N|} + \sum_{i=1}^{|C|} C_i$, where $C_i$ are the discrete domains of each categorical feature.

% For fair comparison across all competitors, and in line with the \emph{CARLA} framework \cite{carla}

\subsection{Transformer-based VAE for tabular data}
\label{sec:architecture}
% Our work extends the architecture described in \tabsyn\ and uses it for CF generation.

As previously mentioned, we build on the architecture used in \cite{tabsyn} for synthetic mixed tabular data generation, and adapt it for CF generation. In particular, we employ learnable tokenizers to process the input data and transformers to learn the latent space. To reconstruct precise one-hot samples while maintaining end-to-end differentiability we propose a Gumbel detokenizer. The entire training pipeline of the VAE is presented in Figure \ref{fig:vae}, and is described in detail hereafter.

% Our work builds on the \tabsyn\ VAE used to generate mixed-type tabular data. We introduce a tokenizer/detokenizer combination and adjust the encoder to generate high-quality CFs. 
% % \tabsyn relies on one encoding Transformer for the $\mu$. We keep the original approach and use an encoder for $\mu$ and $\sigma$. 
% Figure \ref{fig:vae} illustrates the training pipeline of the VAE. The outer boxes represent one sample with mixed feature types; the left shows an original training sample $x$, and the right represents the reconstructed sample $\hat{x}$. The tokenizer multiplies $d$ learnable weights and biases after min-max-normalizing the numerical features and one-hot-encoding the categorical. The detokenizer processes the VAE's output to yield a reconstructed sample in input space. This section discusses each part of the model pipeline in detail.

\noindent{\textbf{Feature Tokenizer.}} To adapt the transformer architecture for tabular data, \cite{tokenizer} proposes a \emph{Feature Tokenizer} as a learnable pre-processing step that converts features into tokens for the subsequent feature-level transformer layers. Specifically, given an input vector $x$ of size $k$,

$x=[x^{num}_1, x^{num}_2, \ldots, x^{num}_{|N|}, x^{oh}_1, x^{oh}_2, \ldots, x^{oh}_{|C|}]$ 

\noindent a tokenized vector $x_{token}$ of size $k \times d$ is created

$x_{token} = [t^{num}_1, t^{num}_2, \ldots, t^{num}_{|N|}, t^{cat}_1, t^{cat}_2, \ldots, t^{cat}_{|C|}]$, 

\noindent via linear transformation of numerical, and "lookup tables" for each categorical feature. In detail we have,

$t^{num} = x^{num} \cdot \mathbf{W}^{num} + \mathbf{b}^{num}$ and 

$t^{cat}_i = x^{oh}_i \cdot \mathbf{W}^{cat}_i + \mathbf{b}^{cat}_i$ for $i \in \{1, \ldots, |C|\}$

\noindent where each column is a token $t_i^{num}, t_i^{cat} \in \mathbb{R}^{1 \times d}$. All weights and biases of the tokenizer, i.e., $\mathbf{W}^{num}, \mathbf{b}^{num} \in \mathbb{R}^{|N| \times d}$ and  $\mathbf{W}^{cat}_i \in \mathbb{R}^{C_i \times d}, \mathbf{b}^{cat}_i \in \mathbb{R}^{1 \times d}$ are learnable parameters.

The learned column-wise token embeddings are passed to the transformer-based encoder that captures the rich feature interdependencies to output the mean and log variance. The latent vector $z = \mu + \sigma \cdot \epsilon$ is obtained via parameterization \cite{vae}. The same inverse procedure is followed for reconstructing the token vector $\hat{x}_{token}$ through the decoder. To get the final output $\hat{x}$ we use our specialized \emph{Gumbel detokenizer}, described hereafter.  

\noindent{\textbf{Gumbel-Detokenizer.}} A key requirement for \us\ is a fully differentiable pipeline that allows for optimizing CF samples in the latent space through the back-propagation of gradients. Additionally, it is essential to ensure that the decoded samples adhere to feature-type constraints. For instance, decoded categorical vectors $\hat{x}^{cat}_i$ should be one-hot. Considering this need to reconstruct one-hot data while enabling gradient flow, we introduce a Gumbel detokenizer that uses the Gumbel-softmax \cite{gumbel} trick to generate categorical features. In particular the $gumbel(.)$ function,  
\[
gumbel(x_i) = \frac{exp((log(x_i)+g_i)/\tau}{\sum^{C_i}_{j=1} exp((log(x_j)+g_j)/\tau},
\]
where $C_i$ denotes the number of modes per categorical feature, $g$ denotes the probabilistic variable sampled from the Gumbel distribution, and $\tau$ denotes the temperature hyperparameter. It is important to note that (for Section~\ref{sec:feature_utilization}) the $gumbel$ function outputs the discretized one-hot samples but uses the \emph{soft samples} for differentiation. After decoding we get the reconstructions, 

% It extends the softmax function to enable a sharper binarization of the categories within each categorical feature by utilizing the reparametrization trick on a probabilistic variable of the Gumbel distribution. 
% The sharper one-hot resemblance compared to the softmax function enables \us 's loss functions to thrive, preventing the need for inferior encodings for categorical features. More specifically, we have:

\[
\hat{x}^{num} = \text{sigmoid}(\hat{t}^{num} \cdot \mathbf{\hat{W}}^{num} + \mathbf{\hat{b}}^{num})
\]
\[
\hat{x}^{oh}_i = \text{gumbel}( \hat{t}^{cat}_i \cdot \mathbf{\hat{W}}^{cat}_i + \mathbf{\hat{b}}^{cat}_i) \text{\ for \ } i \in \{1, \ldots, |C|\}
\] 

This ensures that all input constraints are respected for the reconstructed samples $\hat{x}=[\hat{x}^{num}_1, \hat{x}^{num}_2, \ldots, \hat{x}^{num}_{|N|}, \hat{x}^{oh}_1, \hat{x}^{oh}_2, \ldots, \hat{x}^{oh}_{|C|}]$, i.e. numerical columns are in the range of $[0,1]$ and that all categorical vectors follow a one-hot distribution. In conclusion, \us\ enables gradient-based CF generation in the latent space and inherently guarantees tabular feature constraints. In contrast, other gradient-based methods resort to postprocessing techniques or rely on additional regularization losses to maintain categorical constraints \cite{carla, dice}. In our experiments, we show that this leads to unwanted feature-type bias. 

\noindent{\textbf{VAE training.}} 
% The resulting transformer-based VAE is trained using the standard $\beta$-VAE loss $\mathcal{L} = \|x-\hat{x}\| + \beta \cdot \mathcal{L}_{KL}$. Following \cite{tabsyn}, we opt for a better reconstruction, than a perfectly Gaussian latent space, by gradually decreasing $\beta = [\beta_{max},\beta_{min}]$ during training. A graphical representation of the entire architecture is presented in Figure \ref{fig:vae}.
We train the transformer-based VAE using the $\beta$-VAE loss, $\mathcal{L} = \|x-\hat{x}\| + \beta \cdot \mathcal{L}_{KL}$ \cite{bvae}, where $\mathcal{L}_{KL}$ denotes the discrete Kullback-Leibler divergence between the latent variable and a standard gaussian.
Following \cite{tabsyn}, we opt for a better reconstruction than a perfectly Gaussian distributed latent space by gradually decreasing $\beta = [\beta_{max},\beta_{min}]$ during training.

\subsection{Counterfactuals in the latent space}
\label{sec:generation}
After training the transformer-based VAE, we use the latent representations to search for CFs by traversing the latent space via gradient steps. We use a loss term comprised of three components designed for \emph{Validity}, \emph{Proximity}, and \emph{Sparsity} (referring to the desiderata in Section \ref{sec:related}). 

More specifically, given an input instance $x_0$, we obtain the initial latent representation $z_0=Enc(x_0)$ through the encoder. We then initialize the optimization with $z=z_0$ and take gradient steps updating $z$ with $\nabla_z \mathcal{L}_{\text{CF}}$, minimizing the following loss function:
\begin{align}
\mathcal{L}_{\text{CF}}(z) = & \phantom{{}+{}} \text{hinge\_yloss}\big[f\Big(Dec(z)\Big), y=1\big] \nonumber \\
& + \lambda_{prox\_input} \cdot \|x_0 - Dec(z)\|_1 \nonumber \\
& + \lambda_{prox\_latent} \cdot \|z_0 - z\|_2
\label{eq:loss}
\end{align}

\noindent \textbf{Validity.} The first component computes the difference between the target class ($y=1$) and the current prediction of the black-box model $f(.)$. The latent vector $z$ being optimized, is first passed through the decoder $Dec(z)$. This reconstructed vector is in the original tabular form, by design of our architecture, and can be directly used to get the prediction of the model $f(Dec(z))$. Following \cite{dice}, we use the hinge-loss, defined as
\[
 \text{hinge\_yloss} = max\big[0, 1-logit\big(f(.)\big)\big]
\]
which has two functionalities, i) it heavily penalizes predictions that do not belong to the target class, i.e. when $\mathbb{P}(f(x) = 1) < 0.5$, and ii) it returns a penalty when the target class is achieved, i.e. when $\mathbb{P}(f(x) = 1) \geq 0.5$, proportional to the difference $\mathbb{P}(f(x) = 1) - 0.5$ between the predicted target probability and the decision threshold. The intended effect is pushing instances across the decision boundary to ensure validity, while optimizing for better trade-off solutions (i.e. proximity and sparsity) afterward.\\
\noindent \textbf{Input proximity and sparsity.} The second term, $\|x_0-Dec(z)\|_1$ measures the L1 distance of the original instance $x_0$ and the reconstructed sample, where $\lambda_{prox\_input}$ is a weighting hyperparameter. This term serves as a \emph{proximity} loss in the input space. Using the L1 norm (instead of e.g. L2) additionally encourages feature sparsity \cite{zhou2023scgan}. It is important to note that although sparsity can be computed using the L0 norm, i.e. count distance, this operation is non-differentiable.\\  
\noindent \textbf{Latent proximity.} The last term, $||z_0-z||_2$ measures the L2 distance to the original latent representation $z_0$, to encourage proximity in the latent space, where $\lambda_{prox\_latent}$ is a weighting hyperparameter.

Motivated by other works, that either contain the search in a neighborhood of the latent space \cite{cchvae}, or minimize the distance in the input space \cite{revise,clue}, we decide to follow a combined approach ensuring both latent and input proximity. Our ablation study (Section \ref{sec:results_lossablation}) empirically demonstrates that this combined method yields superior results in terms of proximity and sparsity. Our optimization process is illustrated in Figure \ref{fig:cf_optimization}.

% Lipschitz stuff

% Standard training loss terms and non-linear activation functions create a non-Lipschitz continuous model function, or a function with large Lipschitz constant. We are not regularizing model parameters during VAE training to mitigate this effect like others \cite{khromov2024fundamentalaspectslipschitzcontinuity, gouk2020regularisationneuralnetworksenforcing, NEURIPS2018_d54e99a6}. Therefore, the latent and input data manifolds might differ in scale and shape, motivating the impact of losses in both the latent and decoded input space. Our loss ablation study in \ref{sec:results_lossablation} supports this observation.
% A function $f: X \rightarrow Y$ is said to be Lipschitz continuous if it satisfies $D_a(f(x_1),f(x_2)) \leq k D_b(x_1, x_2) \forall x_1,x_2 \in X$, for a real-valued $k \geq 0$, and metrics $D_a$ and $D_b$. 

% $\forall x,y \in \mathbb{R}^n, ||f(x)-f(y)||_{2} \leq ||x-y||_{2}$

% Notably, by design of the decoder and our Gumbel-detokenizer, reconstructed samples $\hat{x} = Dec(z)$ adhere to the input space constraints that are used during training of $f(.)$, eliminating the need for any additional post-processing to perform a forward pass.

% \begin{figure}[h]
%   \centering
%   \includegraphics[width=\linewidth]{figures/Sample.png}
%   \caption{Overview of the counterfactual generation process.}
%   \label{fig:cf_optimization}
% \end{figure}

\section{Experimental evaluation}
\label{sec:experiments}
This section presents the datasets, baselines, metrics, and the general setup of our quantitative evaluation. 

\subsection{Experimental setup}
We initialize \us\ by training the VAE for $4.000$ epochs, gradually decreasing $\beta$ from $\beta_{max}=10^{-3}$ to $\beta_{min}=10^{-5}$, and use $\tau=1.0$ for the Gumbel distribution. We choose a maximum of $30.000$ training samples for larger datasets to improve comparability. After the VAE is sufficiently trained, we perform Stochastic Gradient Descent (SGD) to find CFs for a maximum of $5.000$ steps or until the loss converges and the target class is reached. For loss weighting, we set $\lambda_{prox\_input}=1$ and $\lambda_{prox\_latent}=1$, as these hyperparameters yield the best overall results. We discuss this further in our ablation study (Section~\ref{sec:results_lossablation}). For all baseline competitors we use the implementations of the CARLA framework \cite{carla}.  

\subsection{Real world financial datasets}
We choose five real-world financial datasets, with a range of different data types, feature counts, and ratios of categorical to numerical features, to ensure a thorough experimental evaluation. In Table~\ref{tab:datasets} we list the characteristics of each dataset, such as the size of the training set used in our experiments, the number of numerical and categorical features, and a description of the binary classification problem.

% Specifically, the classification target for the \lending, \gmc, and \default\ \cite{frank2010uci} datasets is the likelihood of full loan repayment, for the \bank\ \cite{frank2010uci} dataset it is the likelihood of subscription, and for the \adult\ \cite{frank2010uci} dataset the objective is the income level prediction.

% OLD LAYOUT TABLE 1
% \begin{table}[h]
% \caption{Datasets overview.}
% \centering
% \begin{tabular}{|c|c|c|c|c|c|}
% \hline
% \textbf{Dataset} & \makecell[c]{\textbf{Training}\\\textbf{Size}} & \makecell[c]{\textbf{\#Features}\\$\mathbf{(Num/Cat)}$} & \textbf{Target Class} \\ \hline
% \makecell[c]{Lending\\Club} & 50.000 & 8/4 & \makecell[c]{Loan status\\\{\textbf{\textcolor{darkgreen}{fully paid}},\\ charged off\}} \\ \hline
% \makecell[c]{Give Me\\Some Credit} & 50.000 & 6/3 & \makecell[c]{SeriousDlqin2yrs\\\{\textbf{\textcolor{darkgreen}{no}}, yes\}} \\ \hline
% \makecell[c]{Bank\\Marketing} & 50.000 & 7/9 & \makecell[c]{Deposit subscription\\\{\textbf{\textcolor{darkgreen}{yes}}, no\}} \\ \hline
% \makecell[c]{Credit\\Default}  & 27.000 & 14/9 & \makecell[c]{Default payment\\\{\textbf{\textcolor{darkgreen}{yes}}, no\}} \\ \hline
% \makecell[c]{Adult\\Census} & 32.000 & 4/8 & \makecell[c]{Income\\\{\textcolor{darkgreen}{$\mathbf{\geq50K}$}, $<50K$\}} \\ \hline
% \makecell[c]{Dutch\\Census} & 50.000 & 0/11 & \makecell[c]{Occupation\\\{\textbf{\textcolor{darkgreen}{high-level}},\\ low-level\}} \\ \hline
% \end{tabular}

% \label{tab:datasets}
% \end{table}

\begin{table}[h]
\caption{Dataset characteristics.}
\label{tab:datasets}
\centering
\begin{tabular}{lccc}
\toprule
\midrule
\textbf{Dataset} & \makecell[c]{\textbf{Training}\\\textbf{Size}} & \makecell[c]{\textbf{\#Features}\\$\mathbf{(Num/Cat)}$} & \textbf{Target Class} \\ \hline
\makecell[l]{Lending\\Club} & 30.000 & 8/4 & \makecell[c]{Loan status\\\{\textbf{\textcolor{darkgreen}{fully paid}},\\ charged off\}} \\ \hline
\makecell[l]{Give Me\\Some Credit} & 30.000 & 6/3 & \makecell[c]{SeriousDlqin2yrs\\\{\textbf{\textcolor{darkgreen}{no}}, yes\}} \\ \hline
\makecell[l]{Bank\\Marketing} & 30.000 & 7/9 & \makecell[c]{Deposit subscription\\\{\textbf{\textcolor{darkgreen}{yes}}, no\}} \\ \hline
\makecell[l]{Credit\\Default}  & 27.000 & 14/9 & \makecell[c]{Default payment\\\{\textbf{\textcolor{darkgreen}{yes}}, no\}} \\ \hline
\makecell[l]{Adult\\Census} & 32.000 & 4/8 & \makecell[c]{Income\\\{\textcolor{darkgreen}{$\mathbf{\geq50K}$}, $<50K$\}} \\ \hline
% \makecell[l]{Dutch\\Census} & 50.000 & 0/11 & \makecell[c]{Occupation\\\{\textbf{\textcolor{darkgreen}{high-level}},\\ low-level\}}\\

\bottomrule
\bottomrule
\end{tabular}
\end{table}

\noindent \textbf{Lending Club.} The Lending Club public dataset \cite{jagtiani2019roles} comprises detailed information on loans issued by the Lending Club company, including borrower characteristics, loan specifics, and performance indicators such as payment status. It serves as a resource for analyzing the financial performance of peer-to-peer loans with payment status as the target.

\noindent \textbf{Give Me Some Credit}. The Give Me Some Credit dataset \cite{cchvae} contains anonymized records of credit users, focusing on features such as their debt-to-income ratio, monthly income, and number of open credit lines. The target variable is a delay in payment for more than 90 days over the last two years.

\noindent \textbf{Bank Marketing.} The Bank Marketing dataset \cite{frank2010uci} contains information from a marketing campaign by a Portuguese bank, including client details, campaign contact information, and the outcome of each contact. It is used for predictive modeling to determine the likelihood of clients subscribing to a term deposit.

\noindent \textbf{Credit Default.} The Credit Default dataset \cite{frank2010uci} includes information on clients' credit card behavior, such as payment history, credit limits, and demographic details. The target is a default in payment.

\noindent \textbf{Adult Census.} The Adult dataset \cite{frank2010uci} contains demographic information and income data from the 1994 U.S. Census. The individual's income level serves as the target. The prediction is based on features like age, occupation, and capital gain.

% \noindent \textbf{Dutch.} The Dutch Census Dataset includes demographic and socioeconomic information from the 2001 Dutch Census, featuring attributes like age, education level, employment status and household composition. The occupation level serves a the target.

\subsection{Baseline competitors}
\label{sec:baselines}

We compare \us\ to related methods on counterfactual generation that either directly optimize in the input space or employ generative models to first learn latent representations (see Section~\ref{sec:related}). The descriptions of all methods are listed below.

\subsubsection{Baselines operating in input space}\

% \begin{enumerate}[label=--, leftmargin=*]
% \item \textbf{Input space}\\
\noindent We compare to two state-of-the-art gradient-based methods that do not leverage latent representations. 
% \begin{enumerate}[label=\textbullet, leftmargin=0pt]
% \item \textbf{\wachter} et al. 

\noindent \textbf{\wachter}\ \cite{wachter} first mathematically defined CF explanations, using Stochastic Gradient Descent (SGD) to optimize a loss function. The objective is twofold, optimizing for the label flip of the black-box model, while minimizing the distance to the original sample. However, this method tends to find CFs very close to the decision boundary \cite{verma2020counterfactual}, and clamps categorical features to their original values \cite{wachter, carla}. 

\noindent \textbf{\dice}\ \cite{dice} extends the previous approach, taking into account more practical considerations regarding the feature types, such as incorporating a regularization loss for enforcing one-hot representations during optimization (see Section~\ref{sec:feature_utilization}). 
% \end{enumerate}

\subsubsection{Baselines operating in latent space} \

\noindent More recent works, similar to our approach, utilize generative autoencoders to learn latent representations of the data, before searching for CFs in that latent space. We have identified two such approaches for VAE-based tabular data CFs.

% \begin{enumerate}[label=\textbullet, leftmargin=0pt]
\noindent \textbf{\revise}\ \cite{revise} employs a VAE of fully connected neural network layers to learn a structured latent space. As with other gradient-based approaches \cite{dice,wachter}, a loss function is optimized to change the predicted class while minimizing the distance to the original instance. This distance is measured in the reconstructed input, similar to our input-proximity loss term. Gradient descent updates the latent representations until the loss converges, after which the final CFs are returned by the decoder.

\noindent  \textbf{\cchvae}\ \cite{cchvae} similarly uses a conditional VAE for representation learning. Unlike earlier methods, it does not optimize a loss; rather CFs are found in the latent space using a model-agnostic search algorithm. Specifically, multiple points are sampled in the latent space with a growing-sphere approach, until some decoded sample matches the CF requirements, i.e., minimal proximity. Although such heuristic-based methods are more efficient, they do not guarantee optimal results and can be more difficult to adapt to new objectives or constraints.

Although the competitors employ various methods to handle categorical data, they all share the common practice of discretization (e.g. rounding) after each optimization step. Our experiments in Section~\ref{sec:feature_utilization} demonstrate that this results in feature-type bias.

% \end{enumerate}
% \end{enumerate}

\subsection{Metrics}

We evaluate \us\ and all baselines along several metrics to assess the effectiveness of each method. To ensure a fair comparison, each metric is calculated directly in the input space. Additionally, we evaluate all methods on the same test set, selecting $n = 1000$, previously unseen, instances that are not part of the target class, i.e., $X^0_{\text{test}} = \{ x^0_i \mid f(x^0_i) = 0, \; i = 1, 2, \ldots, n \}$. Each method generates a set of \emph{valid} counterfactuals $CF_{\text{test}} = \{ x'_i \mid f(x'_i) = 1\}$. Because CF generation is a difficult non-convex problem \cite{dice}, it is not guaranteed that a \emph{valid} CF will be found for each instance. Therefore, the number of valid CFs, $n_{val} = |CF_{\text{test}}|$, might be less than the number of test instances, i.e., $n_{val} \leq n$. We define all metrics in detail hereafter.

\noindent \textbf{Validity.} The validity score is the most important metric, as it measures the success rate, i.e. the percentage of instances for which optimization successfully switched the decision of the black-box classifier to the target class. More formally, 
\[
\text{Validity ($\uparrow)$} = \frac{n_{val}}{n} = \frac{1}{n} \sum_{i=1}^{n} \mathbb{I}(f(x'_i) = 1)
\] 
Each of the following metrics is calculated only for valid CFs.\\ 
\noindent \textbf{Sparsity.} The sparsity scores measures the percentage of features changed to achieve a CF. We specifically differentiate between categorical and numerical sparsity, 
\[
\text{Sparsity Cat ($\downarrow)$} = \frac{1}{n_{val}} \sum_{i=1}^{n_{val}} \frac{\| x^{0}_{cat} - x'_{cat} \|_0}{|C|}
\] 
\[
\text{Sparsity Num ($\downarrow)$} = \frac{1}{n_{val}} \sum_{i=1}^{n_{val}} \frac{\| x^{0}_{num} - x'_{num} \|_0}{|N|}
\] 
Where the $L0$ norm $\| x^0- x'\|_0$, counts the number of features that have different values in $x^0$ compared to $x'$. The result is normalized by the total number of features, i.e. $|N|$ for numerical and for $|C|$ categorical.\\ 
\noindent \textbf{Proximity.} Proximity uses a distance function to measure how close the CF is to the original instance. It is defined only for numerical features since the categorical sparsity metric essentially plays the role of the proximity metric for categorical features.

\[
\text{Proximity Num ($\downarrow)$} = \frac{1}{n_{val}} \sum_{i=1}^{n_{val}} \|x^0_{num} - x'_{num}\|_1
\] 

The numerical features are standard-normalized, and the L1 norm is used to measure the distance.

% The Gower's distance between two instances is defined as,
% \[
% d_{Gower} (x, \tilde{x}) = \frac{1}{|N|+|C|} \Big[\sum_{i=1}^{|N|} \frac{\|x^{num}_i - \tilde{x}^{num}_i\|_1}{R_i} + \| x^{cat} - \tilde{x}^{cat} \|_0  \Big]
% \]

% Confidence metric not included anymore
% \noindent \textbf{Confidence.} The confidence score measures the probability, assigned by the black-box model $f$, that the CF belongs to the target class. 
% \[
% \text{Confidence ($\uparrow)$} = \frac{1}{n_{val}} \sum_{i=1}^{n_{val}} \mathbb{P}(f(x'_i) = 1)
% \] 

% A score near $0.5$ indicates that the model is almost equally uncertain whether the CF belongs to the target class or not. This lack of confidence suggests that the CF lies near the decision boundary, where small perturbations in input could easily shift back the classification outcome. Larger confidence scores indicate more robust CF explanations.

\subsection{Feature importance and utilization}
\label{sec:feature_importance}

In our feature utilization experiment, we aim to examine the different features altered by each method during CF generation. The general assumption motivating this study is that features with a positive impact on the model prediction (target class) are good candidates for modification, towards a potential label flip. This is especially true if the positive impact comes with a minimal difference to the original value. Such features in the XAI domain, are referred to as \emph{important} features, and corresponding scores, given a black-box model, can be estimated via \emph{feature importance} methods \cite{shap}. We compute feature importance for $X^{0}_{test}$ for the black-box model to study whether essential features are primarily subject to change for \us\ and the competitors.

% For our feature utilization experiment of Section~\ref{sec:feature_utilization}, we want to study the criteria with which the CF generation methods choose which features to change. 

For this study, we choose Shapley Additive Explanations (SHAP) \cite{shap} as the feature importance baseline. SHAP is a field-tested method for ML-based modeling in the finance domain \cite{shapstock}, even further for modeling credit scoring systems \cite{misheva2021explainableaicreditrisk, bussmannShap}. SHAP, originating from cooperative game theory, provides a way to distribute the payoff among players fairly based on their contributions. When applied to neural networks to explain feature importance, SHAP attributes the model's output to its input features by considering all possible feature combinations.

\section{Results and Discussion}
\label{sec:results}
In this section, we discuss the results of our quantitative evaluation. Then we examine how the processing of tabular data by competitors can cause feature-type bias. Finally, we present the results of our ablation study on the hyperparameters of our loss functions.

\subsection{Results for all baselines and datasets}

We evaluate all methods on the five tabular datasets and report the average metric values across the test set. Detailed results for each dataset are presented in Table~\ref{tab:results}. For a comprehensive overview, we further average the results across all datasets to rank the methods, as shown in Table~\ref{tab:overall_results}.

\begin{table}[ht]
  \caption{Results averaged over all datasets.}
  \label{tab:overall_results}
  \setlength{\tabcolsep}{2pt} % Adjust column padding locally
  
  \begingroup
  \begin{tabular}{l c c c c c}
    \toprule
    \midrule
     & \textbf{Validity} (\%) $\uparrow$ & \multicolumn{2}{c}{\textbf{Sparsity}} (\%) $\downarrow$ & \textbf{Proximity} $\downarrow$& \textbf{Top 2} $\uparrow$\\
     & & \textbf{Cat} & \textbf{Num} & \textbf{Num} & \\ 
     \midrule

     \wachter & 0.92 & - & 1.0 & 4.18 & 1\\ 
     \dice & 0.93 & \textbf{0.18} & 0.85 & 2.12 & 12\\ 
     \revise & 0.54 & \underline{0.25} & 0.98 & 2.44 & 3\\ 
     \cchvae & 0.97 & 0.29 & 1.0 & \textbf{0.43} & 12\\ 
     \us\ (us)& \textbf{0.99} & 0.27 & \textbf{0.83} & \underline{1.17} & \textbf{14}\\

  \bottomrule
  \bottomrule
  \end{tabular}
  \endgroup
\end{table}

The results in Table~\ref{tab:overall_results} demonstrate that our method, \us, outperforms the competition. We achieve an almost perfect validity score, finding CFs for $99\%$ of the input instances. Furthermore, in the last column (Top 2), we report the number of times each method ranks first or second best in the per-dataset results of Table~\ref{tab:results}. Here, \us\ stands out, ranking in the top two, for 14 out of a possible 20 times ($70\%$). 

Regarding sparsity and proximity metrics, \us\ changes more categorical features, resulting in $11\%$ worse categorical sparsity on average, compared to the competitors. However, \us\ uses $15\%$ fewer numerical features, while performing minimal changes when they are used. In particular, our performance in numerical proximity ranks second best and is $96\%$ better than competitors on average. We compute these average percentages by comparing the individual performances to \us. 

Overall, compared to the competitors, \us\ finds valid CFs 99\% of the time while making fewer and smaller changes to features on average. Notably, the sparsity metrics reveal a clear bias among all competitors toward using numerical features instead of categorical ones. Specifically, \wachter, \revise, and \cchvae\ alter every single numerical feature when generating counterfactuals, for nearly all test instances (Sparsity Num $= 1$). Similarly, \dice\ utilizes the least categorical features among all methods. We further investigate this observation in Section~\ref{sec:feature_utilization}.

\begin{table}[ht]
  \caption{Results per dataset.}
  \label{tab:results}
  \setlength{\tabcolsep}{4pt} % Adjust column padding locally
  
  \begingroup
  \begin{tabular}{l c c c c}
    \toprule
    \midrule
     & \textbf{Validity} (\%) $\uparrow$ & \multicolumn{2}{c}{\textbf{Sparsity}} (\%) $\downarrow$ & \textbf{Proximity} $\downarrow$\\
     & & \textbf{Cat} & \textbf{Num} & \textbf{Num}\\
     
    \midrule
    \multicolumn{5}{c}{\cellcolor{black!15}\lending}\\
    \midrule
    
     \wachter & 0.95 & - & 1.0 & 1.26 \\ 
     \dice & \underline{0.99} & 0.18 & \textbf{0.85} & 0.83\\ 
     \revise & 0.88 & \textbf{0.07} & 0.99 & 0.80\\ 
     \cchvae & 0.95 & \underline{0.12} & 1.0 & \textbf{0.46}\\ 
     \us & \textbf{1.0} & 0.27 & \underline{0.90} & \underline{0.49}\\
     
    \midrule
    \multicolumn{5}{c}{\cellcolor{black!15}\gmc}\\
    \midrule
     
     \wachter & 0.94 & - & 1.0 & 10.15\\
    \dice & \underline{0.98} & \textbf{0.19} & \underline{0.89} & 3.78\\
    \revise\ & 0.13 & 0.44 & 1.0 & 7.56\\
    \cchvae & 0.96 & \underline{0.61} & 1.0 & \textbf{0.23}\\
    \us & \textbf{1.0} & \underline{0.61} & \textbf{0.86} & \underline{0.42}\\

    \midrule
    \multicolumn{5}{c}{\cellcolor{black!15}\bank}\\
    \midrule

     \wachter & 0.76 & - & 1.0 & 3.87\\ 
     \dice & 0.87 & \textbf{0.19} & \textbf{0.85} & 2.49\\ 
     \revise & 0.51 & 0.25 & 0.94 & \underline{1.86}\\ 
     \cchvae & \underline{0.97} & \underline{0.23} & 1.0 & \textbf{0.46}\\ 
     \us & \textbf{1.0} & 0.35 & \underline{0.89} & 3.69\\ 

    \midrule
    \multicolumn{5}{c}{\cellcolor{black!15}\default}\\
    \midrule

    \wachter & 0.94 & - & 1.0 & 4.57\\
    \dice & 0.98 & \underline{0.31} & \textbf{0.90} & \underline{0.58}\\
    \revise\ & 0.34 & 0.29 & 1.0 & 1.05\\
    \cchvae & \underline{0.99} & \textbf{0.27} & 1.0 & \textbf{0.40}\\
    \us & \textbf{1.0} & 0.38 & \underline{0.99} & \underline{0.58}\\

    \midrule
    \multicolumn{5}{c}{\cellcolor{black!15}\adult}\\
    \midrule
    
     \wachter & \textbf{0.99} & - & 1.0 & 1.06\\
     \dice & 0.85 & \textbf{0.02} & \underline{0.78} & 2.93\\
     \revise & 0.84 & \underline{0.19} & 0.99 & 0.97\\
     \cchvae & \textbf{0.99} & 0.23 & 1.0 & \textbf{0.61}\\
     \us & \underline{0.95} & 0.33 & \textbf{0.54} & \underline{0.67}\\

  \bottomrule
  \bottomrule
  
  \end{tabular}
  \endgroup
\end{table}

\subsection{Feature utilization}
\label{sec:feature_utilization}

Our observations indicate that competitors exhibit bias towards numerical features when identifying CFs, rather than using categorical features. This bias is problematic because an effective method should use features based solely on their influence on finding CFs, irrespective of their type. 

\begin{figure}[ht]
  \centering
  \includegraphics[width=\linewidth]{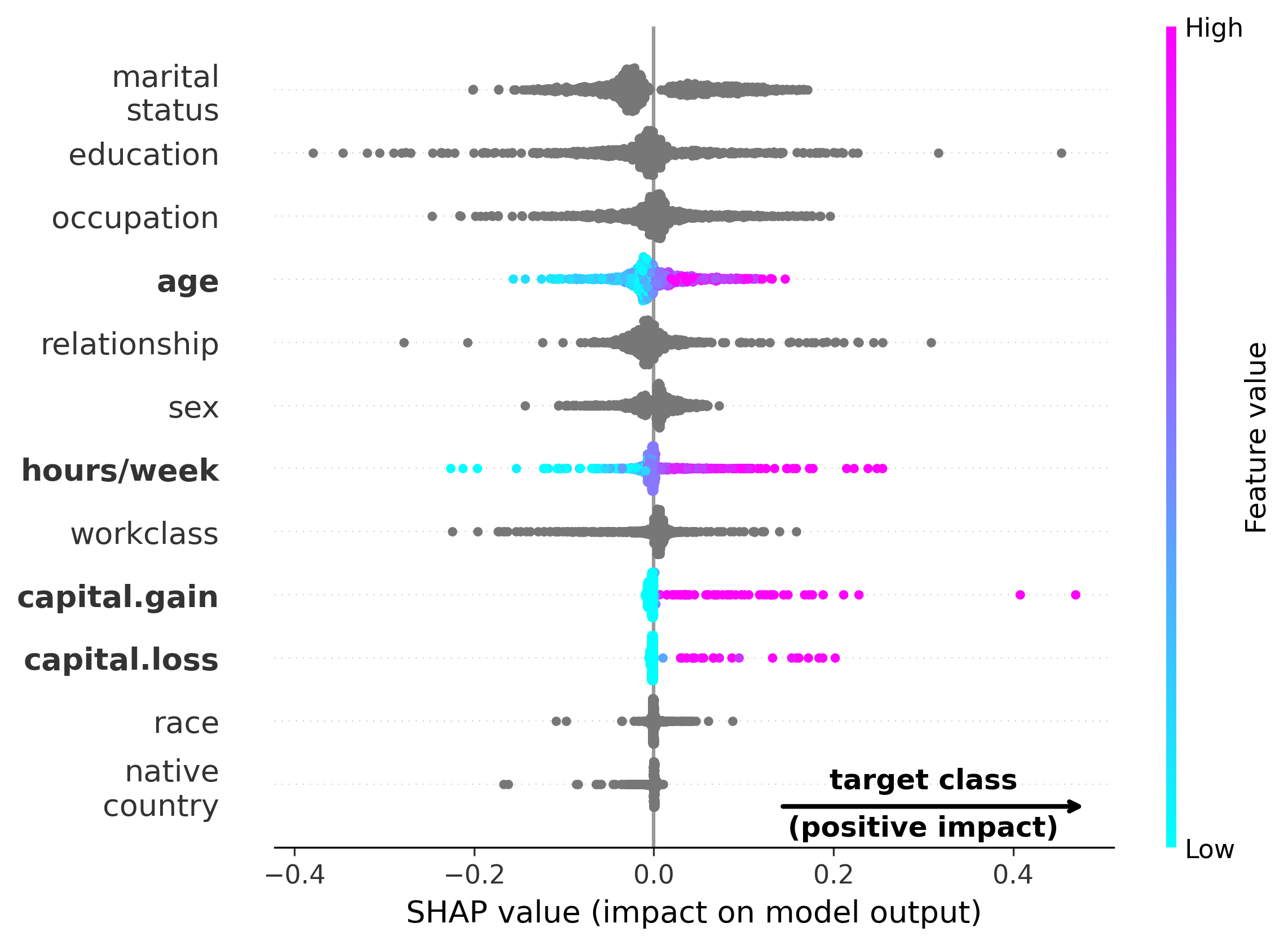}
  \caption{SHAP values for $X^{0}_{test}$ on the \adult\ dataset. Colored dots indicate numerical features and grey dots indicate categorical features. The x-axis displays a positive target class impact to the right and a negative impact to the left.}
  \label{fig:shap}
  \Description{SHAP values for adult.}
\end{figure}

To measure the \emph{importance} of features on the model output, we can use the well-established Shapley explanation method \cite{shap} (as detailed in Section~\ref{sec:feature_importance}). For example, in Figure \ref{fig:shap}, we display the impact of various features from the \adult\ dataset on the output of the black-box classifier. The visualization reveals that categorical features such as education and occupation positively affect the model’s predictions in some cases. Moreover, certain numerical features, like age and hours/week, can have a positive impact even with moderate changes in the feature value (indicated in purple). On the other hand, the capital gain/loss features only show a positive impact when their values are maximal (highlighted in pink). Given that CF methods aim to identify minimal changes with positive outcomes, we expect these methods to favor categorical features, like education and occupation, or numerical features like age and hours/week, when generating CFs for the \adult\ dataset.

\begin{figure*}[ht]
  \centering
  \includegraphics[width=0.9\linewidth]{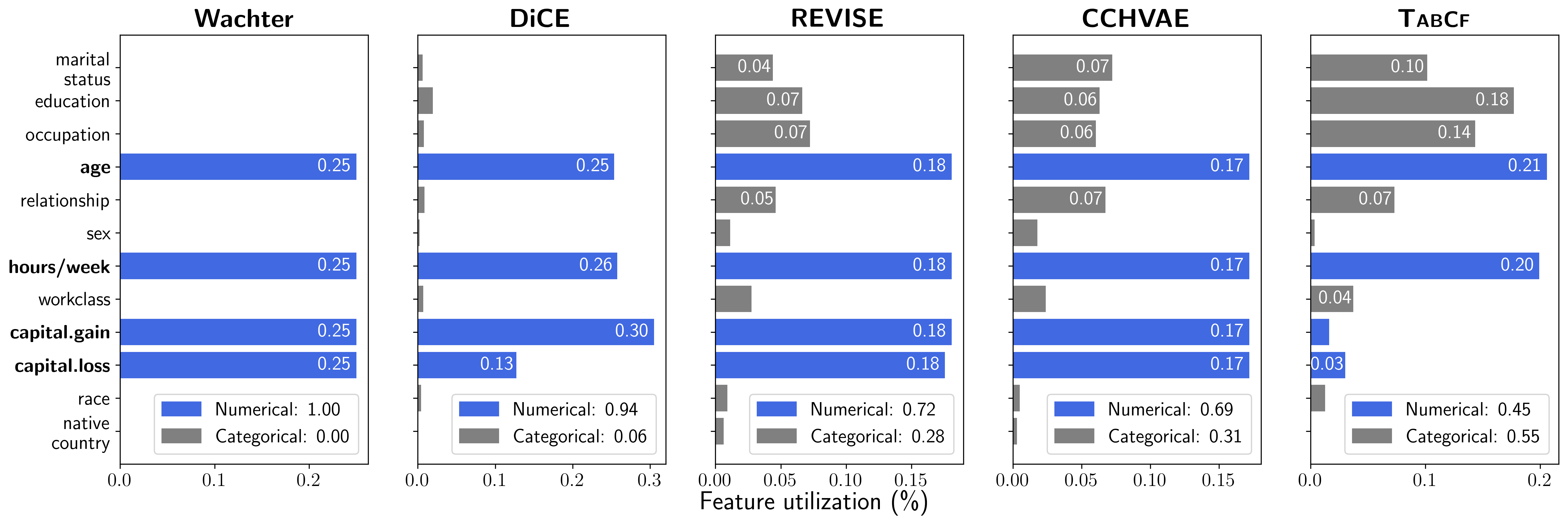}
  \caption{Histogram plot visualizing feature utilization for all methods on the \adult\ dataset. Competitors exhibit feature-type bias, more frequently using numerical features (blue) than categorical features (grey). In contrast, \us\ employs both feature types with similar frequency.}
  \label{fig:feature_changes}
  \Description{Feature changes for the Adult dataset.}
\end{figure*}

However, we empirically show that this is not the case, by computing the \emph{feature utilization} of each method on the \adult\ dataset. The resulting histogram in Figure~\ref{fig:feature_changes} reveals that all competitors primarily use numerical features when generating CFs. \dice\ exhibits a predominant imbalance towards numerical features, utilizing them $94\%$ of the time. While the VAE-based methods (\revise\ and \cchvae) perform slightly better, they still exhibit a strong bias towards using the continuous capital gain/loss features. Our method \us\ stands out as the only one achieving a balanced use of both categorical and numerical features. Our assumption is that this tendency is related to how competitors handle categorical features. Specifically, all baseline implementations discretize categorical columns after each optimization step \cite{carla}. We illustrate why this processing approach can be problematic, with a real example for \dice.

\begin{figure}[ht]
  \centering
  \includegraphics[width=\linewidth]{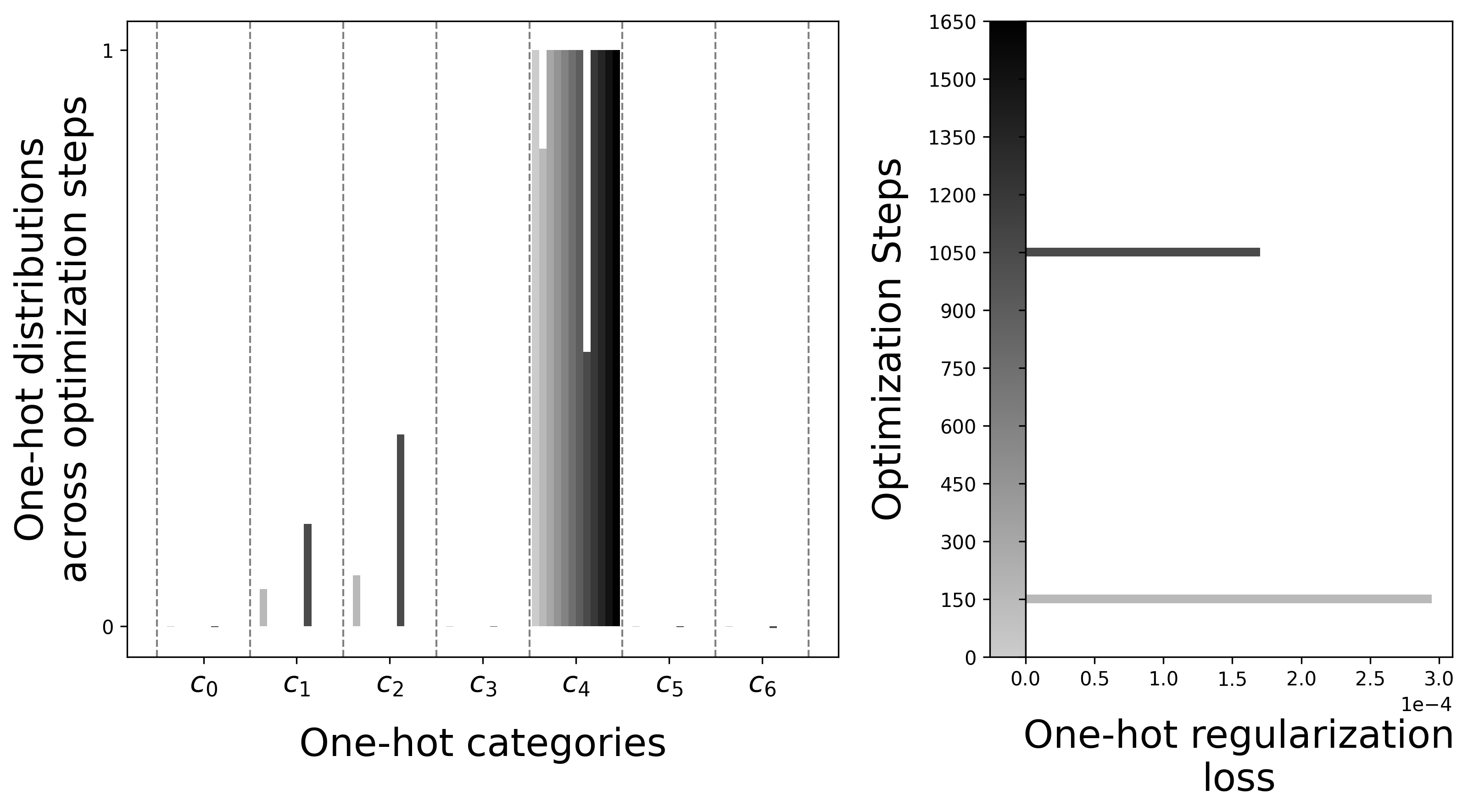}
  \caption{The left plot visualizes a one-hot vector with seven categories ($c_i$ on the x-axis) throughout the optimization process. On the right, the values of the one-hot regularization loss (used by \dice) are plotted (x-axis) over the number of optimization steps (y-axis).}
  \label{fig:dice_loss_example}
  \Description{Loss example for DICE.}
\end{figure}

The \dice\ method calculates a regularization loss, defined as $\mathcal{L}_{reg} = \| \sum[x^{oh}] - 1 \|_2$, for each one-hot encoded vector during optimization. This loss term penalizes vectors $x^{oh}$ that do not sum to 1. Consequently, a gradient update that alters the distribution of $x^{oh}$ results in a positive loss. Additionally, discretization is performed after each optimization step. Figure \ref{fig:dice_loss_example} visualizes the optimization process, showing how the distribution of a one-hot vector $x^{oh} = [c_0, c_1, \ldots, c_6]$ evolves over gradient steps. In two scenarios where a gradient update attempts to change the "hot" value of the feature, the regularization loss (depicted in the right plot) is triggered. Additionally, the gradient updates are insufficient to change the feature value, as the subsequent discretization step always selects the largest value among all $c_i$ to be the "hot" one.

Thus, the regularization loss, together with the discretization approach used by all competitors, introduces feature-type bias towards continuous features. Our method, \us\, avoids this issue by utilizing our Gumbel decoder that produces discretized one-hot reconstructions for querying the black-box model, while using the soft samples for gradient optimization (refer to Section~\ref{sec:architecture}).

\begin{figure}[ht]
  \centering
  \includegraphics[width=\linewidth]{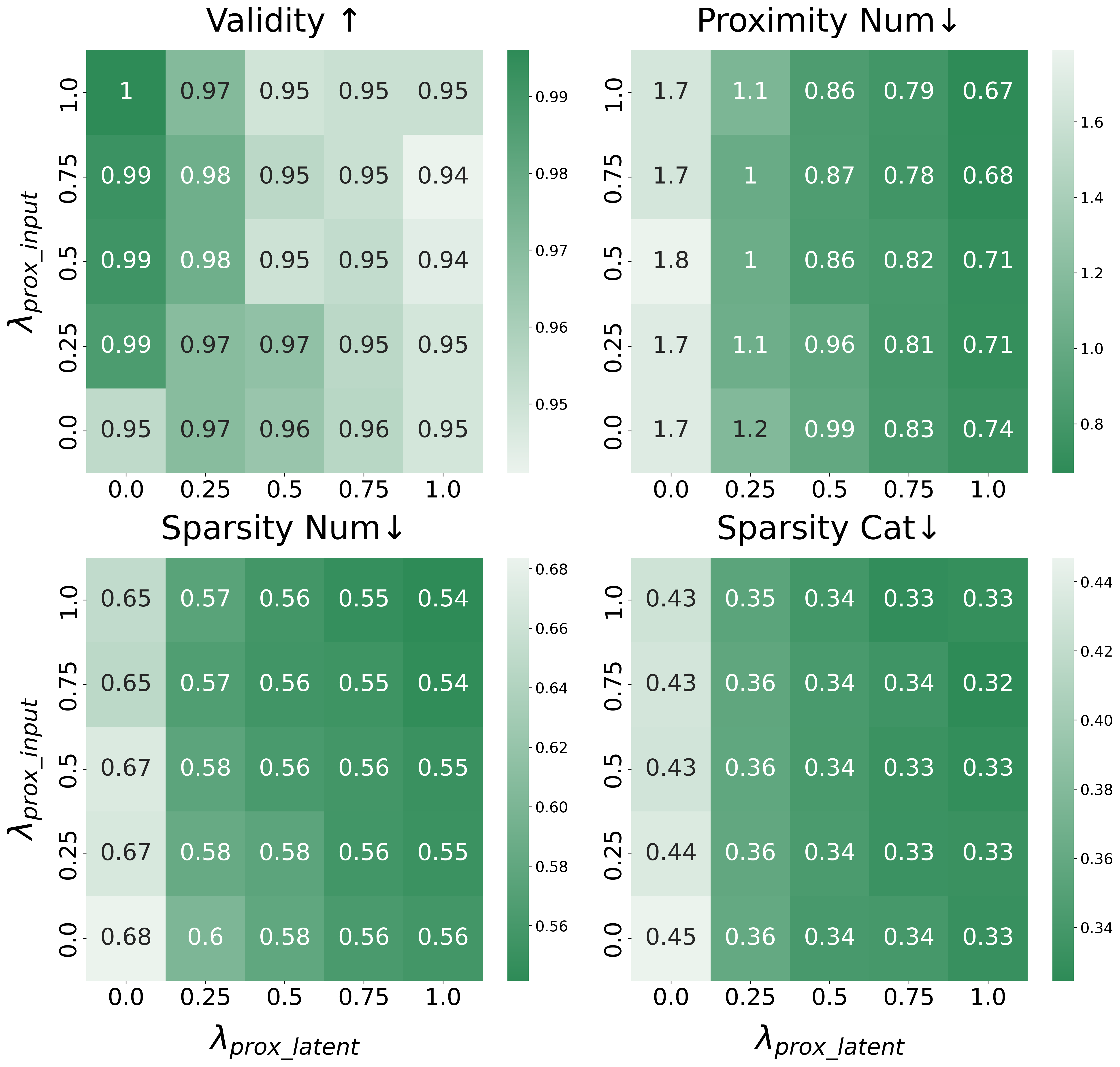}
  \caption{Ablation study for loss hyperparameters on the \adult\ dataset, for the Validity, Proximity, and Sparsity metrics. The x-axes show increasing values for $\lambda_{prox\_latent}$, the y-axes increasing values for $\lambda_{prox\_input}$. A stronger saturation indicates a better score.}
  \label{fig:ablation}
  \Description{Ablation study for loss weightings.}
\end{figure}

\subsection{Ablation losses}
\label{sec:results_lossablation}

The ablation study includes a five-step weight increase in the range [0,1] for $\lambda_{prox\_latent}$ and $\lambda_{prox\_input}$. Therefore, we conduct 25 runs in total, measuring the metrics of Validity, numerical Proximity, and numerical and categorical Sparsity, on each weight combination, for the \adult\ dataset.

As anticipated, we observe the conflicting nature of the validity desideratum, to proximity and sparsity. This trade-off arises because the validity loss term aims to push instances toward the target class, while the proximity loss terms (in the latent and input space), work to keep the CFs close to the original instance. Hence, sparsity and proximity metrics show better scores (more saturated in the plot) for larger values of the hyperparameters, since the loss terms have more influence. As previously discussed, based on the observations of this experiment, we select $\lambda_{prox\_input}=1$ and $\lambda_{prox\_latent}=1$ for weighting both loss terms, in all previous experiments, as these values provided the best overall trade-off.

% \subsection{Immutable features experiment}
% \manolis{TODO: Not done yet}

\section{Conclusion and future work}
\label{sec:conclusion}

This paper presents \emph{\us}, a method that leverages a transformer-based VAE for generating CF explanations for mixed tabular data. Our differentiable Gumbel-Softmax architecture allows precise reconstruction, overcoming feature-type bias present in previous approaches. Additionally, \us\ outperforms competitors in generating valid, proximal, and sparse counterfactuals, thus enhancing the interpretability of black-box models in real-world applications.

In future work, we would like to address user input constraints, such as immutable features or causal relationships between features, which could be achieved by conditioning the latent space. Furthermore, we plan to investigate the effect of distance-preserving Lipschitz-continuous VAEs on proximal counterfactual generation.

\begin{acks}
Our work is funded by the Deutsche Forschungsgemeinsschaft (DFG, German Research Foundation) - SFB1463 - 434502799. We further acknowledge the support by the European Union Horizon Europe Project STELAR, Grant Agreement ID: 101070122.
\end{acks}

% \clearpage

\bibliographystyle{ACM-Reference-Format}
\bibliography{main}

% \appendix

\end{document}